\documentclass[letterpaper, 10 pt, conference]{ieeeconf}  % Comment this line out if you need a4paper

\IEEEoverridecommandlockouts                              % This command is only needed if 
\usepackage{graphicx}
\usepackage{amsmath} % assumes amsmath package installed
\usepackage{amssymb}  % assumes amsmath package installed

\usepackage{amsthm}
\usepackage{bm}
\usepackage{xcolor}
\usepackage{float}
\usepackage{algorithm}
\usepackage{algpseudocode}
\usepackage[colorinlistoftodos]{todonotes}
\usepackage{svg}
\usepackage{dblfloatfix}        % For control over double column images
\makeatletter
\let\NAT@parse\undefined
\makeatother
\usepackage[colorlinks=true, linkcolor=black, citecolor=black, urlcolor=black]{hyperref}
\usepackage{cite}
\usepackage[normalem]{ulem}

\newtheorem{definition}{Definition}

\title{\LARGE \bf
Disentangling Coordiante Frames for Task Specific Motion Retargeting in Teleoperation using Shared Control and VR Controllers
}

\author{Max Grobbel$^{1}$, Daniel Flögel$^{1}$, Philipp Rigoll$^{1}$, Sören Hohmann$^{2}$% <-this % stops a space
\thanks{$^{1}$Max Grobbel, Daniel Flögel and Philipp Rigoll are with FZI - Forschungszentrum Informatik, 76135 Karlsruhe, Germany
        {\tt\small grobbel@fzi.de}}%
\thanks{$^{2}$Sören Hohmann is with the Department of Electrical Engineering, Karlsruhe Institute of Technology, Karlsruhe, Germany}
}

\begin{document}

\maketitle
\thispagestyle{empty}
\pagestyle{empty}

%%%%%%%%%%%%%%%%%%%%%%%%%%%%%%%%%%%%%%%%%%%%%%%%%%%%%%%%%%%%%%%%%%%%%%%%%%%%%%%%
\begin{abstract}

Task performance in terms of task completion time in teleoperation is still far behind compared to humans conducting tasks directly. One large identified impact on this is the human capability to perform transformations and alignments, which is directly influenced by the point of view and the motion retargeting strategy. In modern teleoperation systems, motion retargeting is usually implemented through a one time calibration or switching modes.

Complex tasks, like concatenated screwing, might be difficult, because the operator has to align (e.g. mirror) rotational and translational input commands. Recent research has shown, that the separation of translation and rotation leads to increased task performance. 
This work proposes a formal motion retargeting method, which separates translational and rotational input commands. This method is then included in a optimal control based trajectory planner and shown to work on a UR5e manipulator. 

% \rr{Separation of reference frames in motion retargeting for translation and rotation, thus allowing for task specific alignments. And providing a clear separation between input side and manipulator side (similar to \cite{dejong2004}}

% \todo[inline]{Combined references}

\end{abstract}

%%%%%%%%%%%%%%%%%%%%%%%%%%%%%%%%%%%%%%%%%%%%%%%%%%%%%%%%%%%%%%%%%%%%%%%%%%%%%%%%
\section{INTRODUCTION}
% Motivation for Teleoperation:
The statement by neuroscientist  Daniel Wolpert, "Movement is the only way you have of affecting the world around you" \cite{wolpert2011}, allows the bold assumption that humans could accomplish all tasks remotely through sufficient teleoperation systems. Current research, though, shows \cite{wu2024} that teleoperation performance is one magnitude worse than a human performing tasks directly in terms of task completion times. 

Teleoperation enables humans to interact with the environment in remote places. It has the potential to overcome a number of challenges, including inaccessibility in hazardous environments, lack of human resources, and the need for precision in certain applications \cite{sheridan2016}.
Fields of application include space exploration, handling hazardous materials, surgery and even elder care.
Another upcoming application for teleoperation is the collection of data for imitation learning \cite{wu2024, scherzinger2023} or preference-based learning \cite{heuvel2025} of robots.

% Figure
\begin{figure}
    \centering
    \includegraphics[width=1\linewidth]{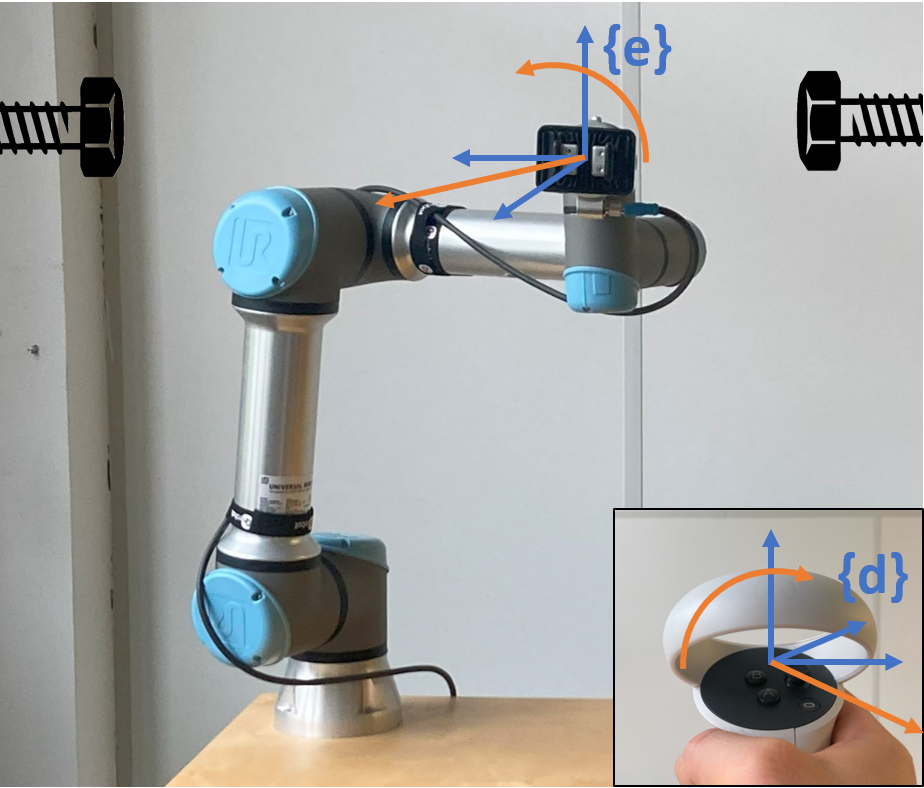}
    \caption{Teleoperated robotic manipulator with input device \{d\}. As the end effector \{e\} is facing the viewpoint of the operator, the operator has to mirror his input commands for translation and rotation.}
    \label{fig:visual-abstract}
\end{figure}

% Problems in teleoperation:
Modern teleoperation systems usually provide haptic feedback to the operator. The additional feedback transports more relevant information to the human operator, leading to higher task performance \cite{wildenbeest2013}, but prices of up to 40,000 \$ \cite{wu2024} make these devices unsuitable for widespread teleoperation applications. In contrast, low-cost input devices like VR (Virtual Reality) controllers or vision-based trackers mitigate the cost problem with prices of just a few hundred dollars, but the reduced feedback to the operator leads to lower performance \cite{wu2024}.

% Need for motion retargeting
Unlocking broad use of teleoperation demands not only affordable input devices, but also intuitive mapping of human motion to robot control commands - commonly referred to as motion retargeting to improve the performance of teleoperation systems. 

This is visualized with the following example. Considering a fixed view frame for the teleoperation scenario, the operator needs to conduct a screwing action to the left and to the right of the robot, e.g. disconnecting or connecting a pipe with a bayonet lock. The operators' input commands must be transformed onto the end effector, which might induce mirrored movements as shown in Fig. \ref{fig:visual-abstract}. This process is referred to as motion retargeting. Recent research \cite{krishnan2022} has shown that separating translational and rotational commands leads to higher performance. In this scenario, it might be more intuitive for the operator to conduct translational movements in the view frame, independent of the orientation of the end effector, while commanding rotational movements in the local frame of the end effector. 

% Insight motion retargeting
Existing work in this field so far defines one reference frame, often called "control frame", in which the operator input will be applied, thus not separating translation and rotation. The choice of this reference frame is usually fixed during a calibration phase \cite{rakita2017}. A formal description of motion retargeting of VR-controller based teleoperation for robotic manipulators is missing, which makes it hard to compare state of art approaches in motion retargeting.

% Insight shared control
A further impact on task performance in teleoperation is the utilization of shared control algorithms \cite{li2023a}. Instead of treating the human inputs in a strict leader–follower manner, shared control decomposes the control tasks between the user (e.g., commanding the end effector pose) and an automated algorithm (e.g., handling joint-level redundancy or collision avoidance) \cite{grobbel2023}. This work leverages model predictive control (MPC) to implement real-time trajectory planning that integrates the user’s end effector commands with the robot’s constraints and objectives.

% Contribution of this paper
The main contribution of this paper is the separation of translation and rotation into distinct reference frames and the introduction of two trees of coordinate frames (the “input device” tree and the “robot manipulator” tree), allowing a clean description of motion retargeting. This enables flexible motion retargeting modes, such as global translation with local rotation, or vice versa which is an extension to previous work \cite{kim2015, hiatt2006}.

This method is embedded into an optimal control-based trajectory planner, offloading the task of inverse kinematics from the human operator to the automation, and shown to work on a UR5e manipulator.

\section{BACKGROUND}
Input devices for teleoperation can be categorized into haptic and non-haptic input devices. A second categorization is the degree of freedoms. Joysticks, for example, only provide two degrees of freedom. This work utilizes a low-cost VR controller as an input device with 6 degrees of freedom. 

\subsection{Motion Retargeting}
Motion retargeting \cite{dragan2013b} - also referred to as motion remapping \cite{rakita2017}, kinematic retargeting \cite{handa2020} or input mapping \cite{grobbel2023} - describes the process of transforming a movement, conducted by the human operator, into a desired motion for a remote-controlled robot. The term motion retargeting is also extended beyond the pure mapping of the desired position, and the control task might be included in the motion retargeting \cite{wang2023}.

The type of motion retargeting depends on the controlled system and the chosen input devices, but a majority of the literature can generally be divided into motion retargeting for humanoid robots \cite{darvish2019, lim2022}, motion retargeting for dexterous hands \cite{li2022, handa2020} and motion retargeting for single manipulators. This paper falls into the later category. 

Motion retargeting for single manipulators can further be divided into a mapping of joint configurations \cite{zhao2016, klein2022}, or a pure mapping of end effector poses (task space control) \cite{rakita2017, praveena2022, lima2024}. The mapping of joint configurations requires kinematically similar input devices or pose estimation of the human operator arm. The human operator is thus enabled to take control of the whole robot directly. This is rather used for humanoid robots and robots with similar kinematics to humans, while "kinematically dissimilar robots are commonly connected at their tips" \cite{siciliano2016}.
Research has shown, that task space control tends to be more intuitive and thus leads to higher teleoperation performance \cite{mower2019, wang2018a}. 

It has been shown that the selection of the control frame impacts the task performance. Humans can only conduct certain transformations between the input movement and the resulting control command \cite{praveena2022}. This can be explained by how humans process different perspectives \cite{taylor1996}. It has also been shown that not only the position of reference frames is relevant, but also the orientation. Humans can only deal with a misalignment of the reference frames up to 30 degrees before performance decreases significantly \cite{kim2015}. 

Recent research \cite{krishnan2022} has shown that the separation of translational and rotational commands leads to higher performance that cannot be achieved with a single reference frame. This motivates the introduction of a motion retargeting method, enabling the separation of translational and rotational commands.

\subsection{Optimal Control based Telerobotics}
Trajectory planning based on optimal control problems is widely used in robot manipulator teleoperation \cite{hu2021, rubagotti2019, selvaggio2022}. Optimal control utilizes model knowledge and an objective function to compute optimal trajectories satisfying system dynamics. The optimal control optimization problem is solved repeatedly in a model predictive control (MPC) manner, resulting in newly planned trajectories.

One challenge in the application of MPC is the real-time capability. With increasing calculation power of modern processors and efficient solvers for optimization problems, MPCs are applied increasingly in robotic applications, e.g., in \cite{faulwassertimm2017}, frequencies of $1 \mathrm{kHz}$ are accomplished, showing that even nonlinear optimization problems are applicable to real-time applications.

In teleoperation, using MPC supports the human operator by including multiple objectives into the objective function or adding further constraints to the optimization problem. For example, by including a minimal distance as constraints, collision-free trajectories can be generated \cite{hu2021, rubagotti2019}. Another example is the stabilization of a liquid in an open container, such as a glass of water. By incorporating a model of the liquid and adding spilling cost to the objective function, slosh-free trajectories can be generated \cite{grobbel2023}. Since the task is \textit{shared} between the human and the automation, those approaches are referred to as \textit{shared control} \cite{li2023a}.

\section{Formal Description of Motion Retargeting with VR controllers in Teleoperation}
\begin{figure*}[h!t]
  \centering
  \framebox{\parbox{0.95\linewidth}{
  \includegraphics[width = 0.8 \linewidth]
  {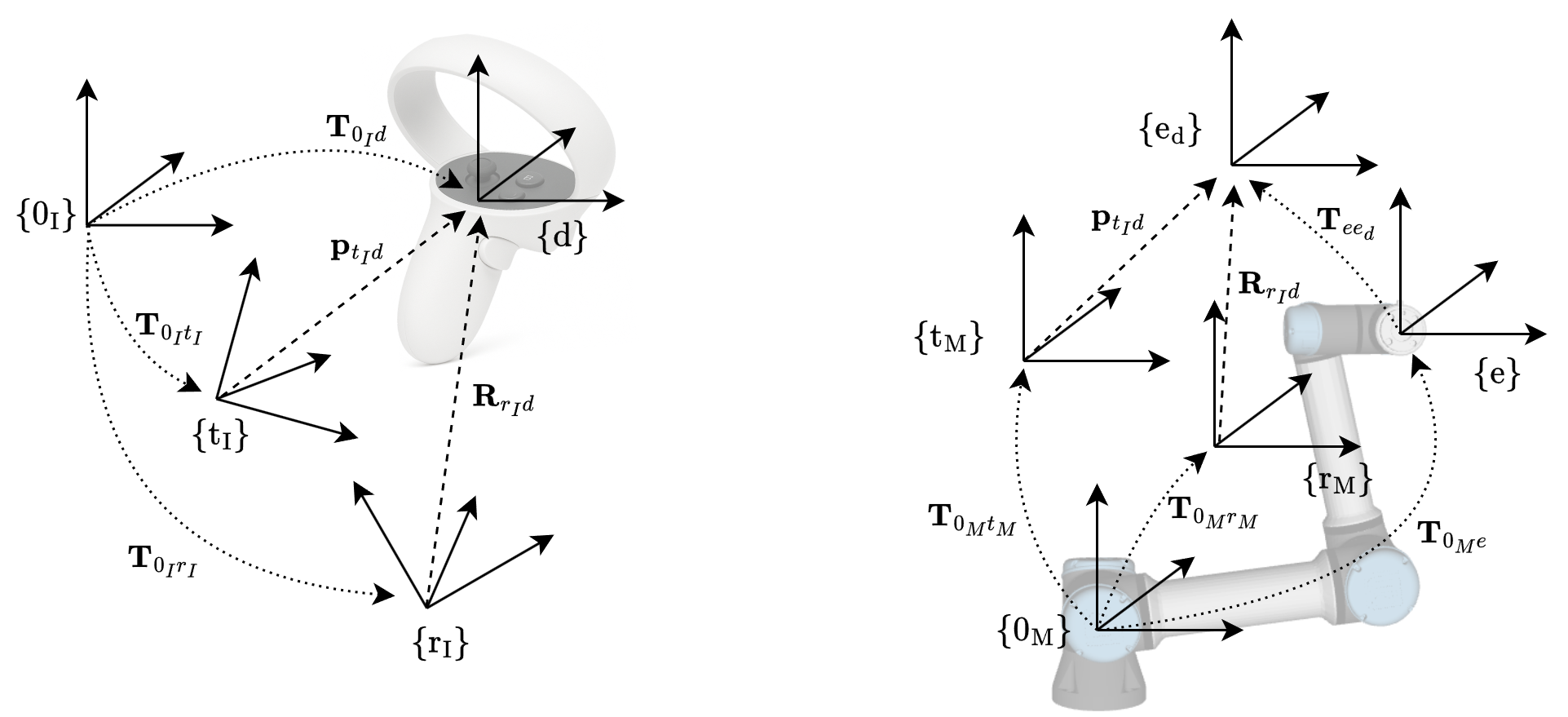}
  \centering
}}
  \caption{Visualization of the two trees of coordinate frames. The tree of frames on the left (input tree) describes the input device with its origin frame \(\{0_I\}\), and the tree on the right (robot tree) represents the robot manipulator with the tool center point frame \(\{e\}\) and its origin frame \(\{0_M\}\). The image of the VR controller was AI-generated.}
  \label{fig:frames}
\end{figure*}

This work introduces a formalized description of motion retargeting using two distinct trees of coordinate frames with separate reference frames for translational and rotational input commands (Fig. \ref{fig:frames}) as explained in  sections \ref{chap:frames} and \ref{chap:retargeting}.

Motion retargeting in the task space, where the operator commands the pose - the combination of position and orientation - of the end effector of a manipulator, can be formally defined as:
\begin{definition}
    \textbf{Motion Retargeting:} The goal of motion retargeting is to enable the human operator to command a desired pose for the end effector of the robot manipulator. 
    The motion retargeting process describes the transformation between two trees of coordinate frames that are not connected. The first tree is the input device space, and the second is the robot space.
\end{definition}

\subsection{Coordinate Frames in Motion Retargeting} \label{chap:frames}
In the following, coordinate frames are written with $\{\}$, and the displacement of coordinate frames is given through homogeneous transformation matrices 
\begin{equation} \label{eq:Transformation}
    \mathbf{T}_{ij} = (\mathbf{R}_{ij}, \mathbf{p}_{ij})
\end{equation}
where $i$ is the origin frame, $j$ is the target frame \cite{lynch2017}, $\mathbf{R}_{ij} \in SO(3)$ is the rotation matrix and $\mathbf{p}_{ij} \in \mathbb{R}^{3}$ the translation vector. Vectors and matrices are depicted with bold letters. Through multiplication of transformation matrices 
\begin{equation}
    \mathbf{T}_{ik} = \mathbf{T}_{ij} \cdot \mathbf{T}_{jk}, 
\end{equation}
the sequential application of transformations can be described as one single transformation. Since the pose of coordinate frames $\{i\}$ is given with a transformation $\mathbf{T}_{0i}$ from the respective origin frame $\{0\}$, transformations starting from $\{0\}$  are used interchangeable with the resulting coordinate frames in the remaining paper.

The two separate trees of coordinate frames for the input device and the robot manipulator are shown in Fig. \ref{fig:frames}. The origin $\{0_I\}$ of the input tree ($I$) can be chosen arbitrarily but is usually given through the software of the input device. The current position and orientation of the input device are represented with the frame $\{d\}$. The measured rotation and translation are given with the transformation matrix $\mathbf{T}_{0_Id}$, starting from the fixed frame $\{0_I\}$. 

The origin $\{0_M\}$ of the tree for the robotic manipulator ($M$) is fixed to the robot's base. The pose $\{e\}$ of the end effector is determined through the kinematic chain of the manipulator. 

For a full separation of translational and rotational input commands, two reference frames on each tree are needed. $\{t_{I}\}$ and $\{t_{M}\}$ are used to track translational displacements on the input device tree and the manipulator tree. $\{r_{I}\}$ and $\{r_{M}\}$ are used to track rotational displacements. Using separate reference frames for translation and rotation allows different strategies of motion retargeting for translation and rotational movements. This allows for task- and user-specific adjustments. For example, mapping the translation can be done in a fixed reference frame while mapping rotational movements in a local reference frame that will be adjusted to the end effector orientation.

\subsection{Motion Retargeting} \label{chap:retargeting}
Teleoperation systems usually contain a clutching mechanism. The operator can activate and deactivate the tracking of movements, preventing unwanted movements from being executed. Two different kinds of motion retargeting can be achieved with the separation of the reference frames, as defined in the following. 

\begin{definition}
    \textbf{Absolute Motion Retargeting:} The two reference frames on the input tree and the two reference frames on the manipulator tree are set to a fixed position during a calibration phase. They won't change until a new calibration is conducted. 
    
\end{definition}
\begin{definition}
    \textbf{Relative Motion Retargeting:} The reference frames on the input tree are changed whenever the clutch is activated. The reference frames on the manipulator tree are reset when the clutch is deactivated.
\end{definition}

\textit{Absolute motion retargeting} also works without the need of a clutch and is the standard choice for humanoid teleoperated robots. It is a good choice for haptic input devices since the alignment of the input device frame with the end effector pose of the robot manipulator is a constant transformation. This mode is comparable to the \textit{view-frame based} control \cite{hiatt2006}.

\textit{Relative motion retargeting} is based on a clutching mechanism. By activating and releasing the clutch, the reference frames can be set to new poses, and only the relative displacements are taken into account for the desired pose of the robot end effector. 

The goal of the motion retargeting process is the calculation of the desired pose of the end effector $\{e_d\}$. This is done distinctly for translation and rotation, as shown in algorithm \ref{alg:motion_retargeting}. 

The relative positional $\mathbf{p}_{t_Id}$ and rotational $\mathbf{R}_{r_Id}$ displacements are taken from the relative transformation matrices in the input tree ($I$) and are applied to the respective reference frames in the manipulator tree ($M$) (algorithm \ref{alg:motion_retargeting}, line \ref{ln:desired_pose}). 
The translational part $\mathbf{p}_{0_{M}e_d}$ of the transformation $\mathbf{T}_{0_Me_d}$ is computed by applying the translation vector 
 $\mathbf{p}_{t_{I}d}$ inside the reference frame $\{t_{M}\}$. This leads to 
 \begin{equation}
     \mathbf{p}_{0_{M}e_d} = \mathbf{p}_{0_Mt_{M}} + \mathbf{R}_{0_Mt_{M}} \cdot \mathbf{p}_{t_{I}d}.
 \end{equation}
Similarly, the rotation matrix $\mathbf{R}_{r_{I}d}$ is applied in the reference frame $\{r_{M}\}$, leading to
 \begin{equation}
     \mathbf{R}_{0_{M}e_d} = \mathbf{R}_{0_{M}r_{M}} \cdot \mathbf{R}_{r_{I}d}.
 \end{equation}
By utilizing eq. \ref{eq:Transformation}, the desired pose for the end effector is given with 
\begin{equation}
    \mathbf{T}_{0_Me_d} = \left( \mathbf{R}_{0_{M}e_d}, \mathbf{p}_{0_{M}e_d}  \right).
\end{equation}

The computation of the input displacements depends on the state of the motion retargeting. The process consists of the active and inactive states. The activation is done by the operator with a clutching mechanism, which is implemented through a button. 

In inactive state, the relative displacements $\mathbf{T}_{t_Id}, \mathbf{T}_{r_Id}$ are set to a unit matrix, thus inducing no relative displacements (algorithm \ref{alg:motion_retargeting}, line \ref{ln:unit_matrix}) in the manipulator tree ($M$). 

In the active state, the frame $\{e_d\}$ follows the operator movements. The displacements on the input tree ($I$) are computed as transformation matrices as shown in algorithm \ref{alg:motion_retargeting}, line \ref{ln:compute_input_displacement}. 

The core concept of the proposed motion retargeting method is the selection of individual orientation matrices for the reference frames. The reference frames on the input tree ($I$) are reset, when the state switches from inactive to active (algorithm \ref{alg:motion_retargeting}, line \ref{ln:switch_active}). For relative motion retargeting, the position of both reference frames is set to the current position of the input device. The orientation of the reference frames is subject to the designer of the system. The reference frames on the manipulator tree ($M$) are reset in the same way, when the state switches from active to inactive (algorithm \ref{alg:motion_retargeting}, line \ref{ln:switch_inactive}), while the selection of the orientations is again subject to the system designer.

The algorithm can easily be extended to \textit{absolute motion retargeting} by also using a calibrated position for the reference frames in either one, or both trees.

\algnewcommand\algorithmicinput{\textbf{Input:}}
\algnewcommand\Input{\item[\algorithmicinput]}
\algnewcommand\algorithmicoutput{\textbf{Output:}}
\algnewcommand\Output{\item[\algorithmicoutput]}

\begin{algorithm}[!h] 
\caption{Relative Motion Retargeting} \label{alg:motion_retargeting}
\begin{algorithmic}[1]
\Input $\mathbf{T}_{0_Id}$, $\mathbf{T}_{0_Me}$, clutch event (activate/deactivate), strategy for choosing rotation matrices
\Output Desired pose for end-effector $\mathbf{T}_{0_Me_d} $
\State Choose $\mathbf{R}_{0_{M}t_{M}}$ 
\State Choose $\mathbf{R}_{0_{M}r_{M}}$
\State \texttt{clutch} $\gets$ \textsc{Off}
\While{true}
  \If{\texttt{clutch} = \textsc{Off}}
    \State Set $\mathbf{T}_{t_Id}, \mathbf{T}_{r_Id} \gets \mathbf{I}$ \quad \# $\mathbf{I}$ unit matrix \label{ln:unit_matrix}
    \If{clutch-event = activate} \label{ln:switch_active}
      \State {Choose $\mathbf{R}_{0_{I}t_{I}}$}
      \State {Choose $\mathbf{R}_{0_{I}r_{I}}$}
      \State {$\mathbf{T}_{0_It_{I}} \gets (\mathbf{p}_{0_Id},\mathbf{R}_{0_{I}t_{I}})$}
      \State {$\mathbf{T}_{0_Ir_{I}} \gets (\mathbf{p}_{0_Id},\mathbf{R}_{0_{I}r_{I}})$}
      
      \State \texttt{clutch} $\gets$ \textsc{On}
    \EndIf
  \ElsIf{\texttt{clutch} = \textsc{On}}

    \State $\mathbf{T}_{t_Id} \gets \mathbf{T}_{0_It_I}^{-1} \cdot \mathbf{T}_{0_Id}$  \label{ln:compute_input_displacement}
    \State $\mathbf{T}_{r_Id} \gets \mathbf{T}_{0_Ir_I}^{-1} \cdot \mathbf{T}_{0_Id}$

    \If{clutch-event = deactivate} \label{ln:switch_inactive}
      \State {Choose $\mathbf{R}_{0_{M}t_{M}}$}
      \State {Choose $\mathbf{R}_{0_{M}r_{M}}$}
      \State $\mathbf{T}_{0_Mt_M} \gets (\mathbf{p}_{0_Me}, \mathbf{R}_{0_Mt_M})$
      \State $\mathbf{T}_{0_Mr_M} \gets (\mathbf{p}_{0_Me}, \mathbf{R}_{0_Mr_M})$
      \State \texttt{clutch} $\gets$ \textsc{Off}
    \EndIf
  \EndIf
  \State \# Set desired end-effector pose:
  \State $\mathbf{R}_{0_Me_d} \gets \mathbf{R}_{0r_{M}} \cdot \mathbf{R}_{r_{I}d}$
  \State $\mathbf{p}_{0_Me_d} \gets \mathbf{p}_{0t_{M}} + \mathbf{R}_{0t_{M}} \cdot \mathbf{p}_{t_{I}d}$
  \State $\mathbf{T}_{0_Me_d} \gets \left(  \mathbf{R}_{0_Me_d}, \mathbf{p}_{0_Me_d} \right)$ \label{ln:desired_pose}
\EndWhile
\end{algorithmic}
\end{algorithm}

\subsection{Mixed Translation and Rotation Retargeting}
To showcase the advantage of the proposed motion retargeting algorithm \ref{alg:motion_retargeting}, a specific selection of rotation matrices is proposed for the initial teleoperation example (Fig. \ref{fig:visual-abstract}). The rotation matrices are selected, such that translational inputs are applied in a calibrated \textit{global} orientation. The rotational inputs are applied in a \textit{local} orientation.

Regardless of the orientation of the input device or the end-effector, the input mapping remains the same. This allows the operator to calibrate the end-effector to his point of view,  without having to mirror further translational inputs. Rotational inputs are further applied in the local frame, such that a clockwise or counterclockwise rotation can be done intuitively with the input device.

This requires a calibration phase, in which the operator selects the orientation of the translational reference frames on the input tree $\{t_I\}$ and the manipulator tree $\{t_M\}$. Those remain constant unless a new calibration is conducted. This allows the operator to align the \textit{global} translational displacements. 

The orientation of the reference frame $\{r_I\}$ is set to the orientation of the input device, when the clutch is activated. Thus, the operator does not have to find the perfect alignment of the input device with respect to its origin $\{0_I\}$.

The orientation of the reference frame $\{r_M\}$ is set to the \textit{upright} end-effector orientation, when the clutch is deactivated. The \textit{upright} orientation is achieved by first rotating about the x-axis, until the y-axis is in the horizontal plane, and then rotation about the y-axis, until the x-axis is in the horizontal plane. The first users found this to be more understandable. 

\section{Shared Optimal Control Trajectory Planning}
\begin{figure*}[!t]
    \centering
    \includegraphics[width=1\linewidth]{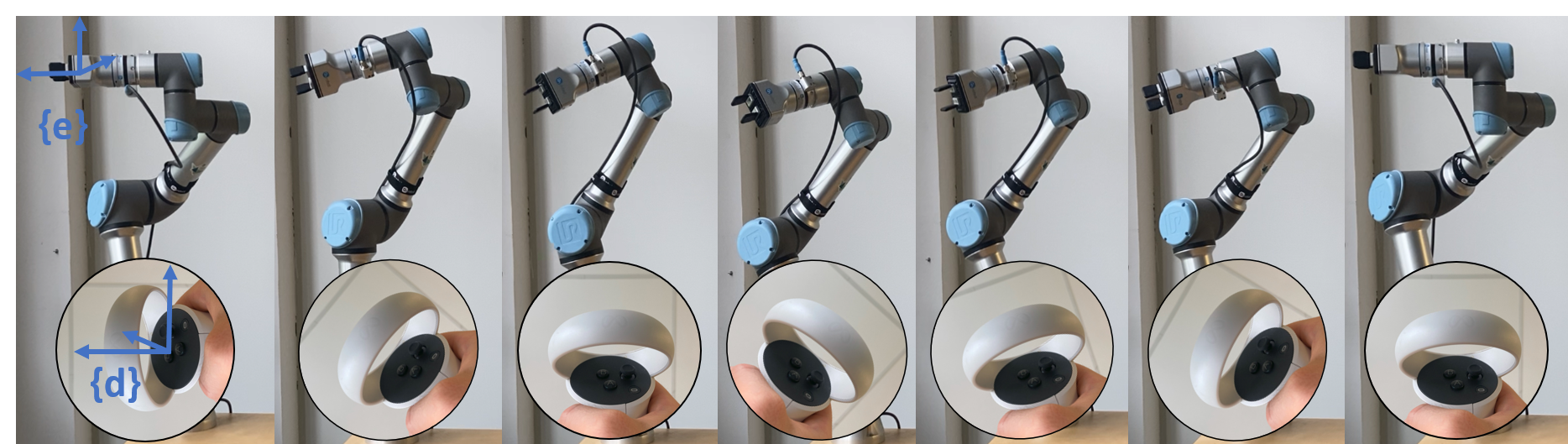}
    \caption{Depiction of a rotational movement command. With the clutching in the first image, the reference frame $\{r_I\}$ is set with the current orientation of the input device $\{d\}$. The orientation for the reference frames $\{t_I\}$ and $\{t_M\}$ was set during a calibration phase. Thus a movement of the VR-controller to the left (from the image perspective) will lead to a movement of the end effector to the left during all stages of this scene.}
    \label{fig:retarget_real_robot}
\end{figure*}

An Optimal Control Problem consists of an objective function $J$, system dynamic constraints, and further constraints on the input and state variables. The objective of the control task is given through the objective function. In this section, the modeling of the robot, the selection of objectives, and the resulting optimization problem for the teleoperation task are described.

\subsection{Robot State Space Modeling}
The existence of low-level controllers is assumed \cite{nubert2020} and \cite{faulwassertimm2017}, such that the joint accelerations serve as control inputs $\mathbf{u} = \mathbf{\ddot{q}}, \mathbf{u} \in \mathbb{R}^{n_q}$ to the considered system. With this assumption, the equations of motion of the manipulator are reduced to a linear dynamic system
\begin{equation} \label{eq:stateSpaceModelRobot}
    \frac{\mathrm{d}}{\mathrm{d}t}
    {\begin{bmatrix}
    \mathbf{q}\\
    \mathbf{{\dot{q}}}
    \end{bmatrix}}
 = \mathbf{A} \begin{bmatrix}
    \mathbf{q}\\ \mathbf{\dot{q}}
    \end{bmatrix} + \mathbf{B} \mathbf{u}
\end{equation}
with 
\begin{align}\nonumber
\mathbf{A} = \begin{bmatrix} \mathbf{0} && \mathbf{\mathbb{I}}\\\mathbf{0} && \mathbf{0} \end{bmatrix}
\qquad \mathrm{and} \qquad
\mathbf{B} = \begin{bmatrix} \mathbf{0} \\ \mathbf{\mathbb{I}} \end{bmatrix},
\end{align}
where $\mathbb{I} \in \mathbb{R}^{n_q \times n_q}$ denotes the identity matrix and $\mathbf{0} \in \mathbb{R}^{n_q \times n_q}$. The state vector $\mathbf{x} = [\mathbf{q}, \mathbf{\dot{q}}]^T$ consists of the concatenation of joint angles and velocities.

A UR5e robot with six rotational joints is used. The joint vector thus becomes {${\mathbf{q} = [q_0,q_1,q_2,q_3,q_4,q_5]}$}. The pose of the end effector is given with the transformation $\mathbf{T}_{0_Me}(\mathbf{q})$ \cite{lynch2017}, which is defined through the forward kinematics using the D-H parameters for the UR5e taken from \cite{UniversalRobotsDH}. In the following, the position of the end effector is denoted with $\mathbf{p}(\mathbf{q})$ and indices are committed. The orientation of the end effector is given with unit quaternions $\alpha(\mathbf{q})$.

\subsection{Human Movement Prediction}
A prediction of the human reference movement is necessary for a predictive control approach. 
Different approaches for human prediction can be used. For example, the findings from \cite{kille2024} might be utilized for a sophisticated trajectory prediction. This work assumes a constant velocity along a straight line over the prediction horizon, leading to a reference of positions $\mathbf{p}_{ref}$. To exclude infeasible reference trajectories, predictions extending over the robotic manipulator's reach are clipped. The reference trajectory of orientation $\alpha_{ref}$ assumes constant orientation.

\subsection{Objective Function}
The main objectives $O_i$ of the trajectory planning task are incorporated in the objective function $J$ of the optimal control problem and defined in the following. An objective function usually consists of stage costs $l_i$ and a final cost $m_i$. 

\paragraph{$O_1$: Tracking}
The main objective of a teleoperation system is to follow the reference pose given by the operator. The prediction of the human motion combined with the motion mapping leads to the reference trajectory $\mathbf{p}_{ref}(t)$ in the manipulator task space.
The squared weighted error is commonly used for position tracking objectives. Thus, the time-dependent stage cost for positional tracking can be defined as 
\begin{equation}
l_{1p}\left(\mathbf{p}(t), \mathbf{p}_{ref}(t)\right) = (\mathbf{p} -\mathbf{p}_{ref})^\mathrm{T} \cdot \mathbf{Q}_p \cdot (\mathbf{p} -\mathbf{p}_{ref})
\end{equation}
with $\mathbf{Q}_p$ as a diagonal matrix containing the elements of the weight vector $\mathbf{w}_{1p} \in \mathbb{R}^3$, assigning individual weights to the three dimensions in the Euclidean space. Note that the stage cost $l_{1p}$ depends on the position $\mathbf{p}$, which depends on the joint configuration $\mathbf{q}$.

Unit quaternions are used for the orientation error, as motivated for task space control in \cite{nakanishi2008operational}. With the definition of a union quaternion 
\begin{equation}
    \mathbf{\alpha} (\mathbf{q}) = \begin{bmatrix}
    \eta & \epsilon_1 & \epsilon_2 & \epsilon_3        
    \end{bmatrix}
\end{equation}
and its components and relation to a rotation of $\phi$ about a unit vector $\mathbf{r}$ with $||\mathbf{r}|| = 1$
\begin{align}
    \eta = \cos(\frac{\phi}{2}) \\
    \epsilon = r \sin (\frac{\phi}{2}) = \begin{bmatrix}
        \epsilon_1 & \epsilon_2 & \epsilon_3
    \end{bmatrix}
\end{align}
which describes the current orientation of the end effector, the orientation error $\mathbf{e} \in \mathbb{R}^3$ is formulated as 
\begin{equation}
    \mathbf{e}(\alpha_d, \alpha(\mathbf{q}) = \eta_d \mathbf{\epsilon}(q) - \eta(q) \mathbf{\epsilon}_d + \mathbf{\epsilon}_d \times \mathbf{\epsilon}(q)
\end{equation}
where the subscript $d$ indicates the desired value and $\times$ is the cross-product. Similar to the positional error, the orientational error is weighted with the diagonal matrix $\mathbf{Q}_o$ with its elements $\mathbf{w}_{1o} \in \mathbb{R}^3$ and reduced to a scalar with 
\begin{equation}
    l_{1o}(\alpha(\mathbf{q}(t)),\alpha_{d}(t)) = \mathbf{e}^T \cdot \mathbf{Q}_o \cdot \mathbf{e}.
\end{equation}

\paragraph{$O_{2}$: Objectives for Stability}
The previously defined objective $O_1$ does not include the joint velocities $\mathbf{\dot{q}}$ and acceleration $\mathbf{\ddot{q}}$. For stability reasons it is recommended to define costs for all system variables and inputs. The quadratic cost term
\begin{align}
    &l_{2a} \left( \mathbf{\dot{q}} \right) = \mathbf{\dot{q}}^T \cdot \mathbf{Q_a} \cdot \mathbf{\dot{q}}\\
    &l_{2b} \left( \mathbf{\ddot{q}} \right) = \mathbf{\ddot{q}}^T \cdot \mathbf{Q_b} \cdot \mathbf{\ddot{q}}
\end{align}
with the diagonal weight matrices $\mathbf{Q_a}$ and $\mathbf{Q_b}$ and the weights $\mathbf{w}_{2a}$ and $\mathbf{w}_{2b}$ for velocity $\mathbf{\dot{q}}$ and acceleration $\mathbf{u}$ is added to the objective function.

The resulting objective function is now given with
\begin{equation}
    J \left( \mathbf{x}(t), \mathbf{u}(t), \mathbf{p}_{ref}(t) \right) = \int_0^T l_{1p} + l_{1o} + l_{2a} + l_{2b} \; \mathrm{d}t,
\end{equation}
where the dependencies on the state trajectories on the right hand side are omitted for readability reasons.

\subsection{Direct Multiple Shooting for Optimal Control}
Combining the system model, the reference trajectory and the objective function with path constraints leads to the optimal control problem
\begin{align} \label{eq:OCP}
    &\underset{\mathbf{x}(t),\mathbf{u}(t)}{\min}  && J \left( \mathbf{x}(t), \mathbf{u}(t), \mathbf{p}_{ref}(t) \right) \\
    &\text{s.t.} \nonumber \\
    &&& \dot{\mathbf{x}} = \mathbf{A} \mathbf{x} + \mathbf{B} \mathbf{u}  \nonumber\\
    &&& \mathbf{q}_{\min} \leq \mathbf{q} (t) \leq \mathbf{q}_{\max} \quad \forall t  \in [0, T] \nonumber\\
    &&& \mathbf{\dot{q}}_{\min} \leq \mathbf{\dot{q}} (t) \leq \mathbf{\dot{q}}_{\max} \quad \forall t \in  [0, T] \nonumber\\
    &&& \mathbf{u}_{\min} \leq \mathbf{u} (t) \leq \mathbf{u}_{\max} \quad \forall t \in  [0, T] \nonumber\\
    &&& \mathbf{x} (t=0) = \mathbf{x}_0 \nonumber.
\end{align}
Even though the system dynamics are linear, the optimal control problem can be assumed to be nonconvex, since the objective function $J$ includes the forward kinematics of the robot manipulator. 

\section{EXPERIMENTAL VALIDATION}
The proposed motion retargeting method is implemented and tested on the scenario as described in the introduction.

\subsection{Hardware and Implementation}
\begin{figure}[ht]
    \centering
    \includegraphics[width=1\linewidth]{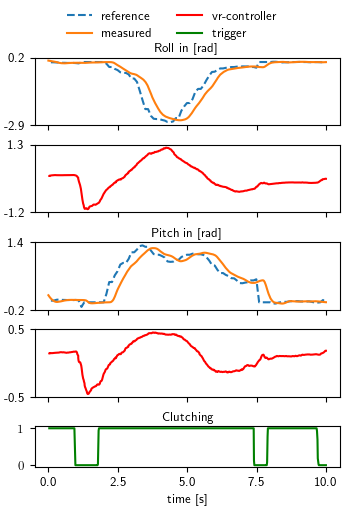}
    \caption{Reference and measured orientation of the end effector and the VR-controller for roll and pitch angles. The bottom graph shows the activation of the clutch button by the operator.}
    \label{fig:validation_teleoperation}
\end{figure}

A Meta Quest 2 VR controller is used as an input device to teleoperate a UR5e robot manipulator. 

The optimal control problem (eq. \ref{eq:OCP}) is solved with the direct multiple shooting method \cite{rawlings2022}. This requires three assumptions or approximations. First, an approximation of the system dynamics with a difference equation. The control input $\mathbf{u}(t)$ is chosen to be a piecewise constant function. Since the system is linear, the exact discretization of the system can be applied. 
The integral part of the objective function is approximated with a Riemann sum, and the path constraints are enforced at the discretization time steps only. 

The resulting nonlinear optimization problem is implemented with Python and CasADi \cite{andersson2019} and solved repeatedly in a model predictive control manner with the interior point solver \textit{IPOPT} \cite{wachter2006} on an \textit{AMD Ryzen 7950X} CPU. The parameters of the objective function are set to $\mathbf{w}_{1p} = \mathbf{w}_{1o} = [100, 100, 100]$ and the diagonal entries of $Q_{2a}$ and $Q_{2b}$ are all set to $0.01$. The prediction horizon is set to $H=10$ steps. 

The resulting control function $\mathbf{u}(t)$ is not utilized directly, but the resulting state trajectories $\mathbf{x}(t)$ are used as a reference for the trajectory tracking controller of the UR5e robot.

\subsection{Initial Results and Discussion}
The motion retargeting is tested on the scenario as described in the introduction. A section of that scenario is a rotational movement (roll) of the end effector (Fig. \ref{fig:retarget_real_robot}) to the left of the point of view. For a larger range of motion, the operator first rotates the VR-controller counter clockwise, before activating the clutch. This allows the operator to conduct a rotation of 180 degrees. Due to the separation of translation and rotation mapping, movements of the VR-controller to the left will lead to movements of the end effector to the left, independent on the current orientation.

The same section is shown in the plots in Fig. \ref{fig:validation_teleoperation}, which shows the setting of the reference frames. The clutch is activated at second 2, with the end effector in horizontal orientation and the VR-controller in a twisted orientation. The rotational movement of the VR-controller in roll and pitch angles leads to reference movements of the end effector. The measurements of the VR-controller are given in the $\{0_I\}$ frame and the orientation of the end effector in the $\{0_M\}$ frame. 

The mismatch between VR-controller movement and the resulting references, shows the importance of motion retargeting. Without a resetting of reference frames, the operator would have to learn the alignment between the VR-controller and the end effector. 

The plots also show a significant pitch movement of the end effector, which was not part of the task. This shows, that even after setting the reference frames, the operator has to learn the alignment of the input device with its respective reference frame \{d\}. This might be supported through a different choice of reference frames. The rotation of the reference frame $\{r_I\}$ could be set to the \textit{upright} orientation of the VR-controller. However, the benefits at the user level remain to be quantified; current results establish technical feasibility.

The calculation time of the optimization problem never exceeded $7 \;$ms during this scenario, indicating a large margin to include further objectives to the optimization problem or implementation on a low-cost CPU. The delay between the reference and the actual position of the end effector is around 0.5 seconds, which is induced through the update rate of $10 \;$Hz of the trajectory planner and a filter on the measurements of the VR-controller.

\section{Conclusion}
A new motion retargeting for teleoperation method, based on four reference frames to separate the mapping of translation and rotation has been introduced. With the formal description through algorithm \ref{alg:motion_retargeting}, this method can easily be implemented to be used with VR-cotrollers or camera based input devices. 

The separation of translation and rotation has been shown to work in an academic example of teleoperation of an UR5e manipulator, including rotational and translational movements. In this work, only one strategy for the selection of orientation of the reference frames has been tested. Comparison of multiple strategies in terms of task performance will be inspected in an upcoming user study.


%%%%%%%%%%%%%%%%%%%%%%%%%%%%%%%%%%%%%%%%%%%%%%%%%%%%%%%%%%%%%%%%%%%%%%%%%%%%%%%%

%%%%%%%%%%%%%%%%%%%%%%%%%%%%%%%%%%%%%%%%%%%%%%%%%%%%%%%%%%%%%%%%%%%%%%%%%%%%%%%%

%%%%%%%%%%%%%%%%%%%%%%%%%%%%%%%%%%%%%%%%%%%%%%%%%%%%%%%%%%%%%%%%%%%%%%%%%%%%%%%%
\bibliographystyle{IEEEtran} % use IEEEtran.bst style
\bibliography{99_references.tex}

\providecommand{\noopsort}[1]{}
\begin{thebibliography}{10}
\providecommand{\url}[1]{#1}
\csname url@rmstyle\endcsname
\providecommand{\newblock}{\relax}
\providecommand{\bibinfo}[2]{#2}
\providecommand\BIBentrySTDinterwordspacing{\spaceskip=0pt\relax}
\providecommand\BIBentryALTinterwordstretchfactor{4}
\providecommand\BIBentryALTinterwordspacing{\spaceskip=\fontdimen2\font plus
\BIBentryALTinterwordstretchfactor\fontdimen3\font minus \fontdimen4\font\relax}
\providecommand\BIBforeignlanguage[2]{{%
\expandafter\ifx\csname l@#1\endcsname\relax
\typeout{** WARNING: IEEEtran.bst: No hyphenation pattern has been}%
\typeout{** loaded for the language `#1'. Using the pattern for}%
\typeout{** the default language instead.}%
\else
\language=\csname l@#1\endcsname
\fi
#2}}

\bibitem{wolpert2011}
\BIBentryALTinterwordspacing
D.~H. Wolpert, ``The real reason for brains,'' TEDGlobal, Edinburgh, Scotland, 2011, accessed 18 Apr. 2025. [Online]. Available: \url{https://www.ted.com/talks/daniel_wolpert_the_real_reason_for_brains}
\BIBentrySTDinterwordspacing

\bibitem{wu2024}
P.~Wu, Y.~Shentu, Z.~Yi, X.~Lin, and P.~Abbeel, ``{{GELLO}}: {{A General}}, {{Low-Cost}}, and {{Intuitive Teleoperation Framework}} for {{Robot Manipulators}},'' in \emph{2024 {{IEEE}}/{{RSJ International Conference}} on {{Intelligent Robots}} and {{Systems}} ({{IROS}})}.\hskip 1em plus 0.5em minus 0.4em\relax IEEE, 2024.

\bibitem{sheridan2016}
T.~B. Sheridan, ``Human-{{Robot Interaction}}: {{Status}} and {{Challenges}},'' \emph{Human Factors}, vol.~58, no.~4, 2016.

\bibitem{scherzinger2023}
S.~Scherzinger, P.~Becker, A.~Roennau, and R.~Dillmann, ``Motion {{Macro Programming}} on {{Assistive Robotic Manipulators}}: {{Three Skill Types}} for {{Everyday Tasks}},'' in \emph{2023 20th {{International Conference}} on {{Ubiquitous Robots}} ({{UR}})}.\hskip 1em plus 0.5em minus 0.4em\relax IEEE, 2023.

\bibitem{heuvel2025}
J.~{\noopsort{heuvel}}de Heuvel, D.~Marta, S.~Holk, I.~Leite, and M.~Bennewitz, ``The {{Impact}} of {{VR}} and {{2D Interfaces}} on {{Human Feedback}} in {{Preference-Based Robot Learning}},'' 2025.

\bibitem{wildenbeest2013}
J.~G.~W. Wildenbeest, D.~A. Abbink, C.~J.~M. Heemskerk, F.~C.~T. {van der Helm}, and H.~Boessenkool, ``The impact of haptic feedback quality on the performance of teleoperated assembly tasks,'' \emph{IEEE Transactions on Haptics}, vol.~6, no.~2, 2013.

\bibitem{krishnan2022}
A.~U. Krishnan, T.-C. Lin, and Z.~Li, ``Design {{Interface Mapping}} for {{Efficient Free-form Tele-manipulation}},'' in \emph{2022 {{IEEE}}/{{RSJ International Conference}} on {{Intelligent Robots}} and {{Systems}} ({{IROS}})}.\hskip 1em plus 0.5em minus 0.4em\relax Kyoto, Japan: IEEE, 2022, pp. 6221--6226.

\bibitem{rakita2017}
D.~Rakita, B.~Mutlu, and M.~Gleicher, ``A {{Motion Retargeting Method}} for {{Effective Mimicry-based Teleoperation}} of {{Robot Arms}},'' in \emph{Proceedings of the 2017 {{ACM}}/{{IEEE International Conference}} on {{Human-Robot Interaction}}}.\hskip 1em plus 0.5em minus 0.4em\relax ACM, 2017.

\bibitem{li2023a}
G.~Li, Q.~Li, C.~Yang, Y.~Su, Z.~Yuan, and X.~Wu, ``The {{Classification}} and {{New Trends}} of {{Shared Control Strategies}} in {{Telerobotic Systems}}: {{A Survey}},'' \emph{IEEE Transactions on Haptics}, vol.~16, no.~2, 2023.

\bibitem{grobbel2023}
M.~Grobbel, B.~Varga, and S.~Hohmann, ``Shared {{Telemanipulation}} with {{VR Controllers}} in an {{Anti Slosh Scenario}},'' in \emph{2023 {{IEEE International Conference}} on {{Systems}}, {{Man}}, and {{Cybernetics}} ({{SMC}})}.\hskip 1em plus 0.5em minus 0.4em\relax IEEE, 2023.

\bibitem{kim2015}
L.~H. Kim, C.~Bargar, Y.~Che, and A.~M. Okamura, ``Effects of master-slave tool misalignment in a teleoperated surgical robot,'' in \emph{2015 {{IEEE International Conference}} on {{Robotics}} and {{Automation}} ({{ICRA}})}, 2015.

\bibitem{hiatt2006}
L.~Hiatt and R.~Simmons, ``Coordinate {{Frames}} in {{Robotic Teleoperation}},'' in \emph{2006 {{IEEE}}/{{RSJ International Conference}} on {{Intelligent Robots}} and {{Systems}}}.\hskip 1em plus 0.5em minus 0.4em\relax IEEE, 2006.

\bibitem{dragan2013b}
A.~D. Dragan, S.~Siddhartha~Srinivasa, and K.~Kenton~Lee, ``Teleoperation with {{Intelligent}} and {{Customizable Interfaces}},'' \emph{Journal of Human-Robot Interaction}, vol.~2, no.~2, 2013.

\bibitem{handa2020}
A.~Handa, K.~Van~Wyk, W.~Yang, J.~Liang, Y.-W. Chao, Q.~Wan, S.~Birchfield, N.~Ratliff, and D.~Fox, ``{{DexPilot}}: {{Vision-Based Teleoperation}} of {{Dexterous Robotic Hand-Arm System}},'' in \emph{2020 {{IEEE International Conference}} on {{Robotics}} and {{Automation}} ({{ICRA}})}.\hskip 1em plus 0.5em minus 0.4em\relax IEEE, 2020.

\bibitem{wang2023}
T.~Wang, H.~Zhang, L.~Chen, D.~Wang, Y.~Wang, and R.~Xiong, ``Robust {{Real-Time Motion Retargeting}} via {{Neural Latent Prediction}},'' in \emph{2023 {{IEEE}}/{{RSJ International Conference}} on {{Intelligent Robots}} and {{Systems}} ({{IROS}})}, 2023.

\bibitem{darvish2019}
K.~Darvish, Y.~Tirupachuri, G.~Romualdi, L.~Rapetti, D.~Ferigo, F.~J.~A. Chavez, and D.~Pucci, ``Whole-{{Body Geometric Retargeting}} for {{Humanoid Robots}},'' in \emph{2019 {{IEEE-RAS}} 19th {{International Conference}} on {{Humanoid Robots}} ({{Humanoids}})}, 2019.

\bibitem{lim2022}
D.~Lim, D.~Kim, and J.~Park, ``Online {{Telemanipulation Framework}} on {{Humanoid}} for both {{Manipulation}} and {{Imitation}},'' in \emph{2022 19th {{International Conference}} on {{Ubiquitous Robots}} ({{UR}})}, 2022.

\bibitem{li2022}
R.~Li, H.~Wang, and Z.~Liu, ``Survey on {{Mapping Human Hand Motion}} to {{Robotic Hands}} for {{Teleoperation}},'' \emph{IEEE Transactions on Circuits and Systems for Video Technology}, vol.~32, no.~5, 2022.

\bibitem{zhao2016}
L.~Zhao, Y.~Liu, K.~Wang, P.~Liang, and R.~Li, ``An intuitive human robot interface for tele-operation,'' in \emph{2016 {{IEEE International Conference}} on {{Real-time Computing}} and {{Robotics}} ({{RCAR}})}.\hskip 1em plus 0.5em minus 0.4em\relax IEEE, 2016.

\bibitem{klein2022}
H.~Klein, N.~Jaquier, A.~Meixner, and T.~Asfour, ``A {{Riemannian Take}} on {{Human Motion Analysis}} and {{Retargeting}},'' in \emph{2022 {{IEEE}}/{{RSJ International Conference}} on {{Intelligent Robots}} and {{Systems}} ({{IROS}})}, 2022.

\bibitem{praveena2022}
P.~Praveena, L.~Molina, Y.~Wang, E.~Senft, B.~Mutlu, and M.~Gleicher, ``Understanding {{Control Frames}} in {{Multi-Camera Robot Telemanipulation}},'' in \emph{2022 17th {{ACM}}/{{IEEE International Conference}} on {{Human-Robot Interaction}} ({{HRI}})}.\hskip 1em plus 0.5em minus 0.4em\relax IEEE, 2022.

\bibitem{lima2024}
R.~Lima, S.~Saha, V.~Vakharia, V.~Vatsal, and K.~Das, ``Augmenting {{Robot Teleoperation}} with {{Shared Autonomy}} via {{Model Predictive Control}},'' in \emph{2024 {{IEEE Conference}} on {{Telepresence}}}.\hskip 1em plus 0.5em minus 0.4em\relax IEEE, 2024.

\bibitem{siciliano2016}
B.~Siciliano and O.~Khatib, \emph{Springer {{Handbook}} of {{Robotics}}}.\hskip 1em plus 0.5em minus 0.4em\relax Springer International Publishing, 2016.

\bibitem{mower2019}
C.~E. Mower, W.~Merkt, A.~Davies, and S.~Vijayakumar, ``Comparing {{Alternate Modes}} of {{Teleoperation}} for {{Constrained Tasks}},'' in \emph{2019 {{IEEE}} 15th {{International Conference}} on {{Automation Science}} and {{Engineering}} ({{CASE}})}.\hskip 1em plus 0.5em minus 0.4em\relax IEEE, 2019.

\bibitem{wang2018a}
Z.~Wang, I.~Reed, and A.~M. Fey, ``Toward {{Intuitive Teleoperation}} in {{Surgery}}: {{Human-Centric Evaluation}} of {{Teleoperation Algorithms}} for {{Robotic Needle Steering}},'' in \emph{2018 {{IEEE International Conference}} on {{Robotics}} and {{Automation}} ({{ICRA}})}.\hskip 1em plus 0.5em minus 0.4em\relax IEEE, 2018.

\bibitem{taylor1996}
H.~A. Taylor and B.~Tversky, ``Perspective in {{Spatial Descriptions}},'' \emph{Journal of Memory and Language}, vol.~35, no.~3, 1996.

\bibitem{hu2021}
S.~Hu, E.~Babaians, M.~Karimi, and E.~Steinbach, ``{{NMPC-MP}}: {{Real-time Nonlinear Model Predictive Control}} for {{Safe Motion Planning}} in {{Manipulator Teleoperation}},'' in \emph{2021 {{IEEE}}/{{RSJ International Conference}} on {{Intelligent Robots}} and {{Systems}} ({{IROS}})}.\hskip 1em plus 0.5em minus 0.4em\relax IEEE, 2021.

\bibitem{rubagotti2019}
M.~Rubagotti, T.~Taunyazov, B.~Omarali, and A.~Shintemirov, ``Semi-autonomous robot teleoperation with obstacle avoidance via model predictive control,'' \emph{IEEE Robotics and Automation Letters}, vol.~4, no.~3, 2019.

\bibitem{selvaggio2022}
M.~Selvaggio, J.~Cacace, C.~Pacchierotti, F.~Ruggiero, and P.~R. Giordano, ``A shared-control teleoperation architecture for nonprehensile object transportation,'' \emph{IEEE Transactions on Robotics}, vol.~38, no.~1, 2022.

\bibitem{faulwassertimm2017}
F.~Timm, W.~Tobias, Z.~Pablo, and F.~Rolf, ``Implementation of {{Nonlinear Model Predictive Path-Following Control}} for an {{Industrial Robot}},'' \emph{IEEE Transactions on Control Systems Technology}, vol.~25, no.~4, 2017.

\bibitem{lynch2017}
K.~M. Lynch and F.~C. Park, \emph{Modern Robotics: {{Mechanics}}, Planning, and Control}.\hskip 1em plus 0.5em minus 0.4em\relax Cambridge University Press, 2017.

\bibitem{nubert2020}
J.~Nubert, J.~Kohler, V.~Berenz, F.~Allg{\"o}wer, and S.~Trimpe, ``Safe and {{Fast Tracking}} on a {{Robot Manipulator}}: {{Robust MPC}} and {{Neural Network Control}},'' \emph{IEEE Robotics and Automation Letters}, vol.~5, no.~2, 2020.

\bibitem{UniversalRobotsDH}
``Universal {{Robots}} - {{DH Parameters}} for calculations of kinematics and dynamics,'' https://www.universal-robots.com/articles/ur/application-installation/dh-parameters-for-calculations-of-kinematics-and-dynamics/, accessed 18 Apr. 2025.

\bibitem{kille2024}
S.~Kille, P.~Leibold, P.~Karg, B.~Varga, and S.~Hohmann, ``Human-{{Variability-Respecting Optimal Control}} for {{Physical Human-Machine Interaction}},'' 2024.

\bibitem{nakanishi2008operational}
J.~Nakanishi, R.~Cory, M.~Mistry, J.~Peters, and S.~Schaal, ``Operational space control: {{A}} theoretical and empirical comparison,'' \emph{The International Journal of Robotics Research}, vol.~27, no.~6, 2008.

\bibitem{rawlings2022}
J.~B. Rawlings, D.~Q. Mayne, and M.~Diehl, \emph{Model Predictive Control: Theory, Computation, and Design}, 2nd~ed.\hskip 1em plus 0.5em minus 0.4em\relax Nob Hill Publishing, 2022.

\bibitem{andersson2019}
J.~A.~E. Andersson, J.~Gillis, G.~Horn, J.~B. Rawlings, and M.~Diehl, ``{{CasADi}}: A software framework for nonlinear optimization and optimal control,'' \emph{Mathematical Programming Computation}, vol.~11, no.~1, 2019.

\bibitem{wachter2006}
A.~W{\"a}chter and L.~T. Biegler, ``On the implementation of an interior-point filter line-search algorithm for large-scale nonlinear programming,'' \emph{Mathematical Programming}, vol. 106, no.~1, 2006.

\end{thebibliography}

\end{document}